\documentclass[]{llncs}
\usepackage{graphicx}
\usepackage[utf8]{inputenc} 
\usepackage[T1]{fontenc}    
\usepackage{hyperref}       
\usepackage{url}            
\usepackage{booktabs}       
\usepackage{algpseudocode}
\usepackage{algorithm}
\usepackage{amsfonts}       
\usepackage{nicefrac}       
\usepackage{microtype}      
\usepackage{xcolor}         
\usepackage{amsmath}
\usepackage{multirow}
\usepackage{lineno}
\usepackage{geometry}
\usepackage{longtable}

\begin{document}
\title{A Short Review and Evaluation of SAM2's Performance in 3D CT Image Segmentation.}

\author{Yufan He\inst{1}, Pengfei Guo\inst{1}, Yucheng Tang\inst{1}, Andriy Myronenko\inst{1}, Vishwesh Nath\inst{1}, Ziyue Xu\inst{1}, Dong Yang\inst{1}, Can Zhao\inst{1}, Daguang Xu\inst{1}, Wenqi Li\inst{1}}
\institute{$^{1}$ NVIDIA}


\maketitle

\begin{abstract}
Since the release of Segment Anything 2 (SAM2), the medical imaging community has been actively evaluating its performance for 3D medical image segmentation. However, different studies have employed varying evaluation pipelines, resulting in conflicting outcomes that obscure a clear understanding of SAM2's capabilities and potential applications. We shortly review existing benchmarks and point out that
the SAM2 paper clearly outlines a zero-shot evaluation pipeline, which simulates user clicks iteratively for up to eight iterations. We reproduced this interactive annotation simulation on 3D CT datasets and provided the results and code~\url{https://github.com/Project-MONAI/VISTA}. Our findings reveal that directly applying SAM2 on 3D medical imaging in a zero-shot manner is far from satisfactory. It is prone to generating false positives when foreground objects disappear, and annotating more slices cannot fully offset this tendency. For smaller single-connected objects like kidney and aorta, SAM2 performs reasonably well but for most organs it is still far behind state-of-the-art 3D annotation methods. More research and innovation are needed for 3D medical imaging community to use SAM2 correctly.
\end{abstract}

\section{Introduction}
The segmentation anything 2~(SAM2)~\cite{ravi2024sam} uses a 2D image encoder with knowledge from previous frames stored in a memory bank. It showed impressive zero-shot 3D understanding and can be naturally applied to 3D medical images like CT and MRI. The research community is debating whether it will replace the well established 3D segmentation methods with 3D operations. We ask three open questions:
\begin{enumerate}
    \item Is the strong zero-shot performance good enough to solve 3D medical image interactive segmentation problem? How far it is compared with state-of-the-art?
    \item Should 3D medical imaging researchers switch to SAM2's model or just use SAM2's dataset? Will 3D-UNet~\cite{nnUNet,myronenko23brats23} type of work achieve better results on 3D medical data if trained on those video dataset? 
    \item Medical image segmentation is usually needed for large-cohort analysis, which requires high-accuracy automatic segmentation models. Are SAM2's architecture and its pretrained weights helpful in producing state-of-the-art automatic segmentation? 
\end{enumerate}

Those questions are important for the community to take next steps. The first step is to evaluate SAM2's performance in 3D interactive medical image segmentation. Several works have reported the benchmark results. However those results showed large variance due to evaluation protocol differences, even for simple organs like liver. 

Ma e.t.c. ~\cite{ma2024segment} initialized the bounding box prompts in the middle slice then the 3D full masks are generated by this \textbf{single box} on this \textbf{single slice}. All the slices without foreground are removed from validation, thus any slice-level false positive are removed. The results showed high variance across organs with lowest dice of 0.14~(inferior vena cava) and highest 0.91~(right kidney). The performance for liver, kidney, spleen and pancreas~(on FLARE22~\cite{ma2023unleashing} with SAM2-Tiny) are approximately 0.58, 0.91, 0.8, 0.17.

Shen e.t.c.~\cite{shen2024interactive} clicked 5 points or used groundtruth mask as prompt to the \textbf{single center slice}. Their observation showed that SAM2 has a large gap with the state-of-the-art methods, The performance for liver, spleen and pancreas~(on MSD~\cite{antonelli2022medical}) are approximately 0.81/0.84, 0.8/0.93, 0.45/0.51. (5 clicks, without/with removing slices that contain no foreground). Both paper showed better results if the center slice is given a mask prompt rather than clicks or box. 

Dong e.t.c.~\cite{dong2024segment} pick \textbf{a single point on a single slices (or three slices uniformly)}. They find that providing 1 point to a single slice for the whole volume segmentation achieve an average performance of 0.36, whereas providing 1 prompt to each slice gives an average performance of 0.50. The performance for liver, kidney, spleen and pancreas are approximately 0.5, 0.2, 0.5, 0.1.~(MSD~\cite{antonelli2022medical} and CT-Org~\cite{rister2020ct}).

Zhu e.t.c.~\cite{zhu2024medical} added prompts to each slice randomly
with a probability of 0.25, their benchmark on BTCV~\cite{fang2020multi} showed a average dice score of 0.5, while the dice for liver, kidney, spleen and pancreas are approximately 0.27, 0.7, 0.36, 0.68.

Although they evaluated the performance on different dataset with different protocol, the variance is too large and conclusion we can draw from them is limited. Major organs~(not diseased) like liver, kidney and spleen should have similar segmentation dice~(usually over 0.9) across different datasets.

In this paper we report a standardized evaluation protocol for SAM2 based on its original paper.

\section{Method}
In the SAM2's evaluation, they used iterative evaluation methods:
\begin{quote}
\textit{We start with click prompts on the first
frame, segment the object throughout the entire video, and then in the next pass, we select the frame with the
lowest segmentation IoU w.r.t. the ground-truth as the new frame for prompting. The model then segments
the object again throughout the video based on all prompts received previously, until reaching a maximum of
Nframe passes (with one new prompted frame in each pass).}
\end{quote}

This is a natural way to mimic user annotation processes, and we follow this general guideline and set the evaluation protocol to be:
\begin{enumerate}
    \item Pick the foreground geometric center as the first positive point and its slice as the starting slice. Get the 2D prediction, then calculate the error region. If false positive region is larger than false negative region, a negative point is selected at the center of false positive region, or else a positive point is selected at the center of false negative region. We select 5 points iteratively to form a good initial segmentation. Note that the error region is eroded by a (3x3) kernel and if there is no pixel left after erosion, the iteration will stop and we will use less than 5 points for the initial segmentation.
    \item The initial slice with 5 clicks will be used to generate the whole 3D scan segmentation with the forward and backward propagation. 2D Dice score and error pixel number is calculated for every slice. The slice with lowest Dice score is selected for the next round of annotation~(if multiple slices have the same lowest dice, use the slice with maximum error pixel number). 
    \item  For the selected slice, we select one positive point from false negative region center, one negative point from false positive region center. If the false positive region size is larger than false negative, a negative point will be randomly selected at the false positive region, and vice versa. We select 3 points in total in accordance with SAM2's protocol, but the exact selection criterion might be different since SAM2 does not reveal the full detail. Meanwhile, the error region is eroded by (3,3) kernel first, so the actual point selected can be smaller than 3. If there is no point selected, the whole process will end. 
    \item The steps 2-3 are repeated 7 times thus we will have maximum 8 edited slices in total. In most cases the total slices are smaller than 8 since they are early stopped due to no point selected in step 3.
    
\end{enumerate}

\section{Results}
\begin{figure}
    \centering
    \includegraphics[width=1\linewidth]{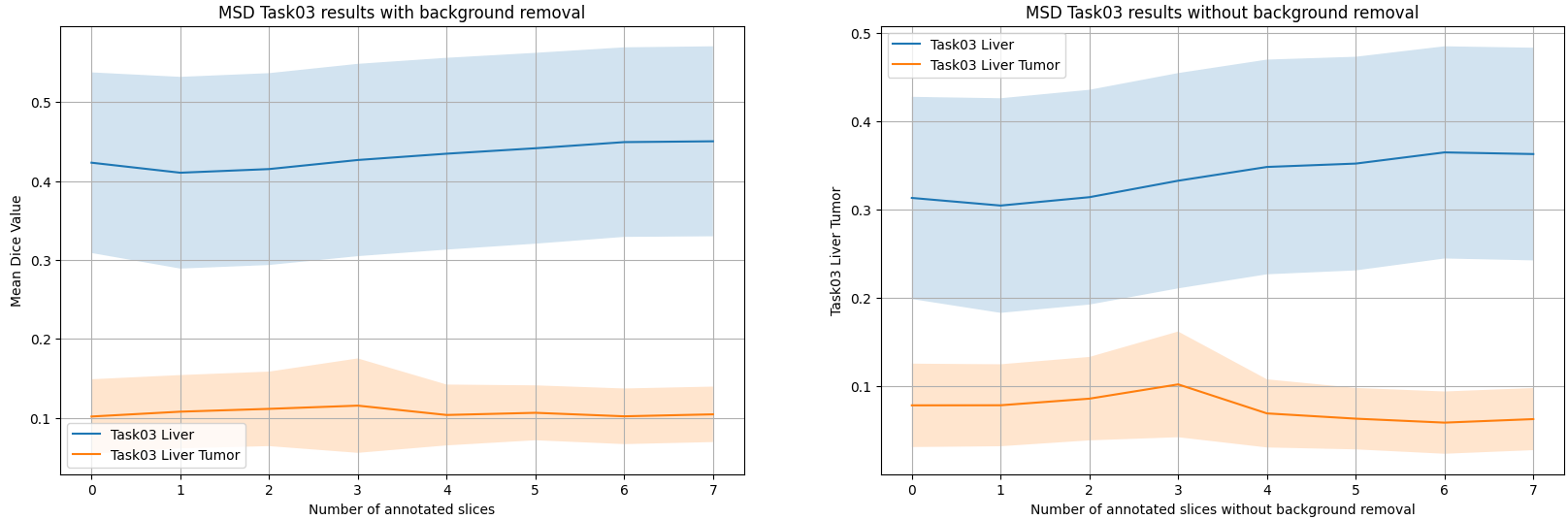} \\
    \includegraphics[width=1\linewidth]{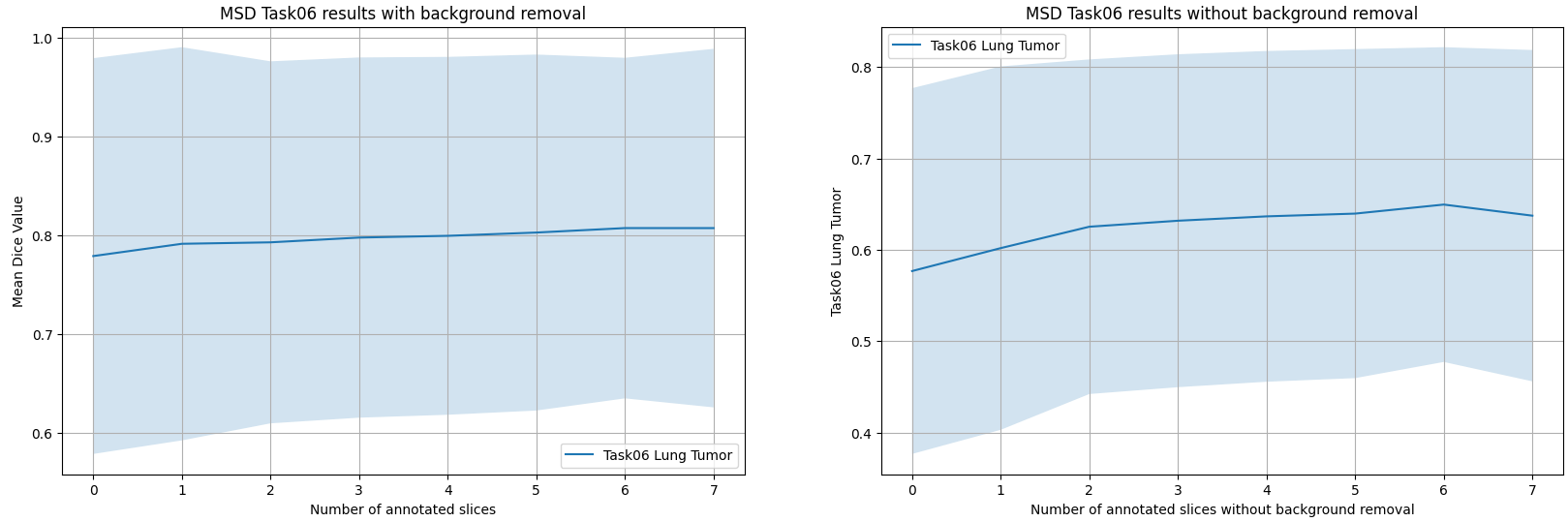} \\
    \includegraphics[width=1\linewidth]{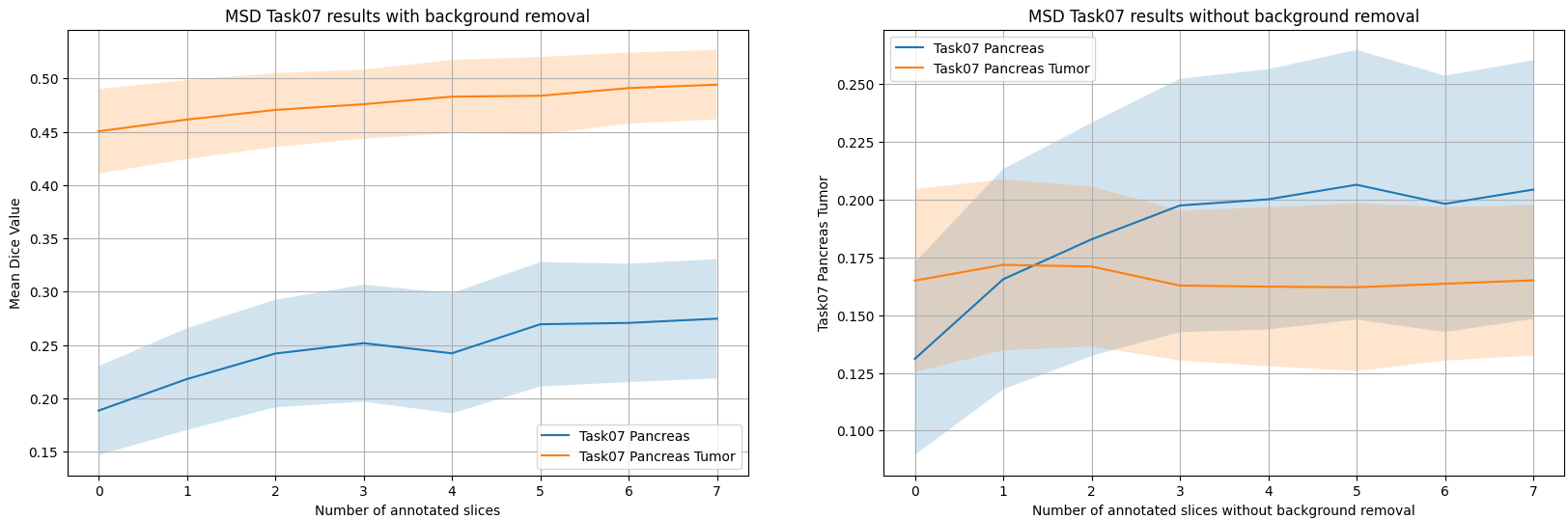} \\
    \includegraphics[width=1\linewidth]{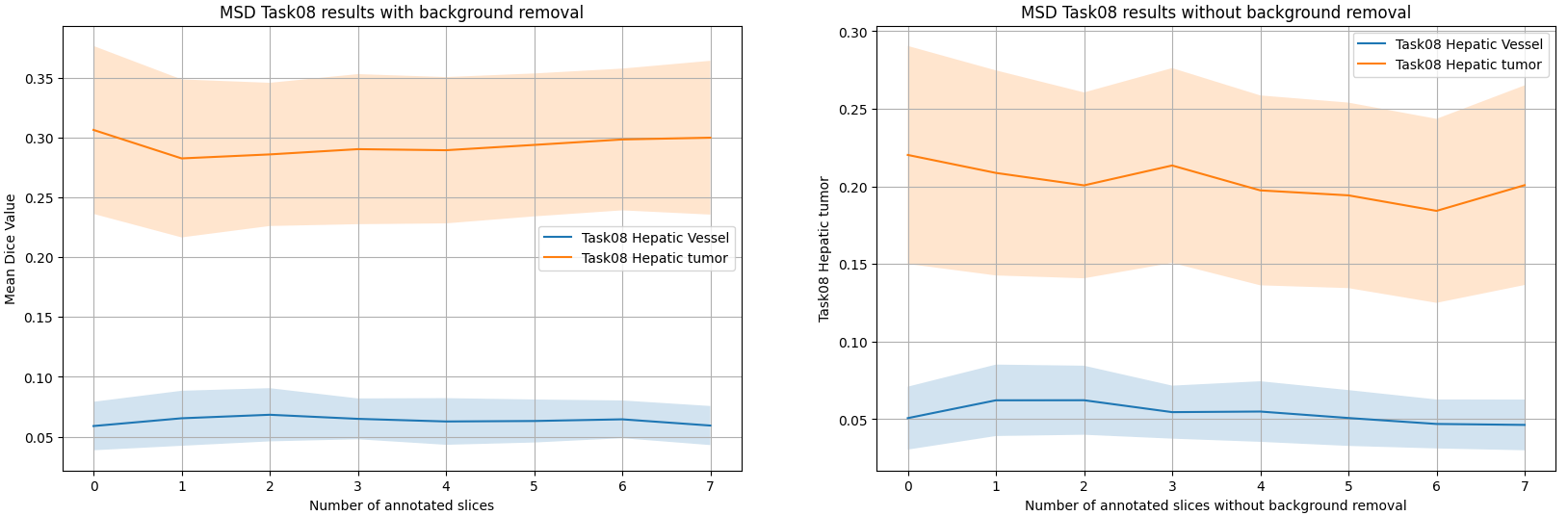} \\
    \caption{Mean dice scores and 95\% confidence interval with annotated slice numbers on MSD tasks}
    \label{fig:plots1}
\end{figure}

\begin{figure}
    \centering
    \includegraphics[width=1\linewidth]{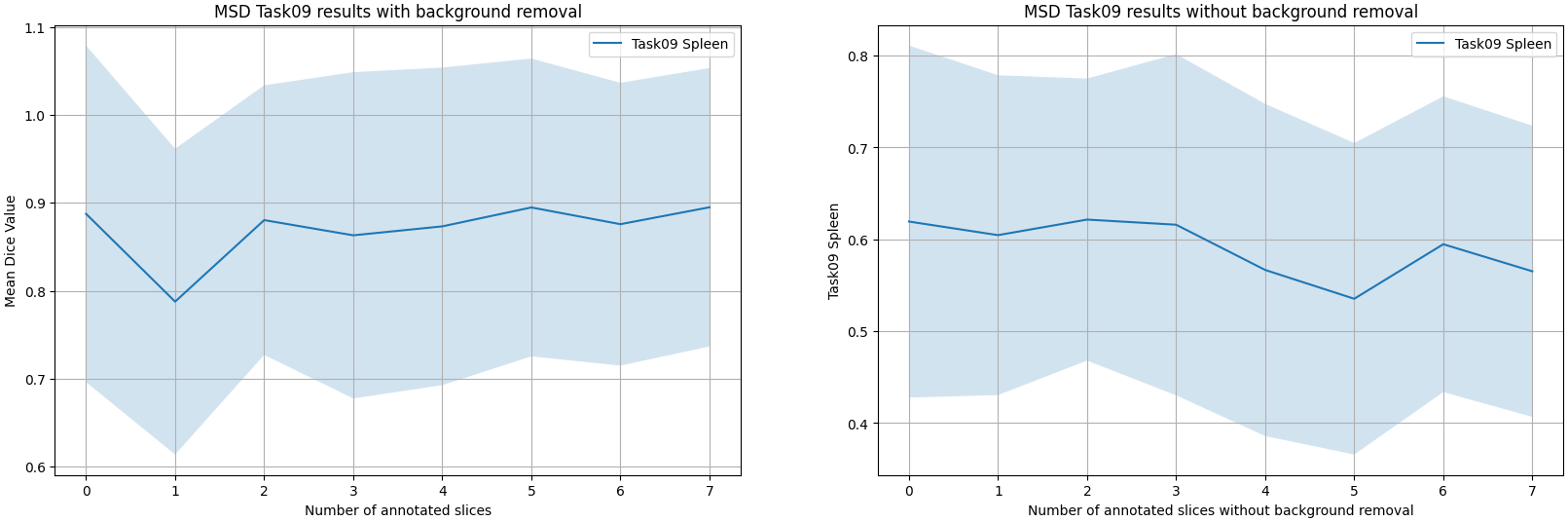} \\
    \includegraphics[width=1\linewidth]{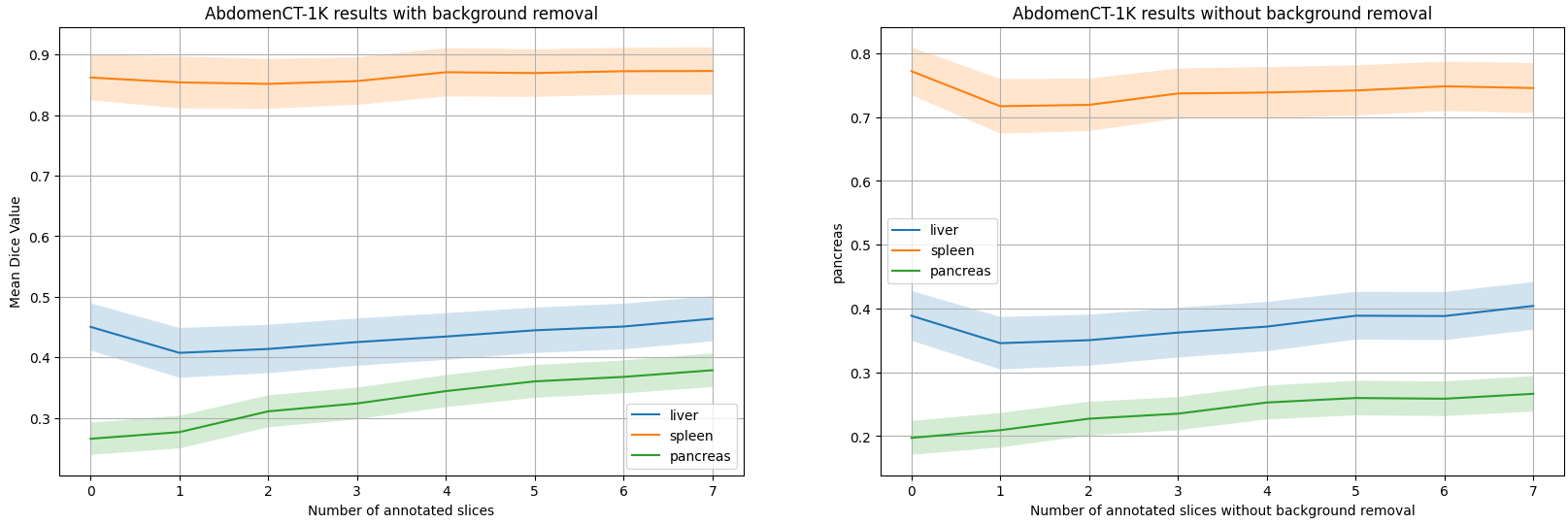} 
    \caption{Mean dice scores and 95\% confidence interval with annotated slice numbers on MSD task09 and AbdomenCT-1K.}
    \label{fig:plots2}
\end{figure}

\begin{figure}
    \centering
    \includegraphics[width=1\linewidth]{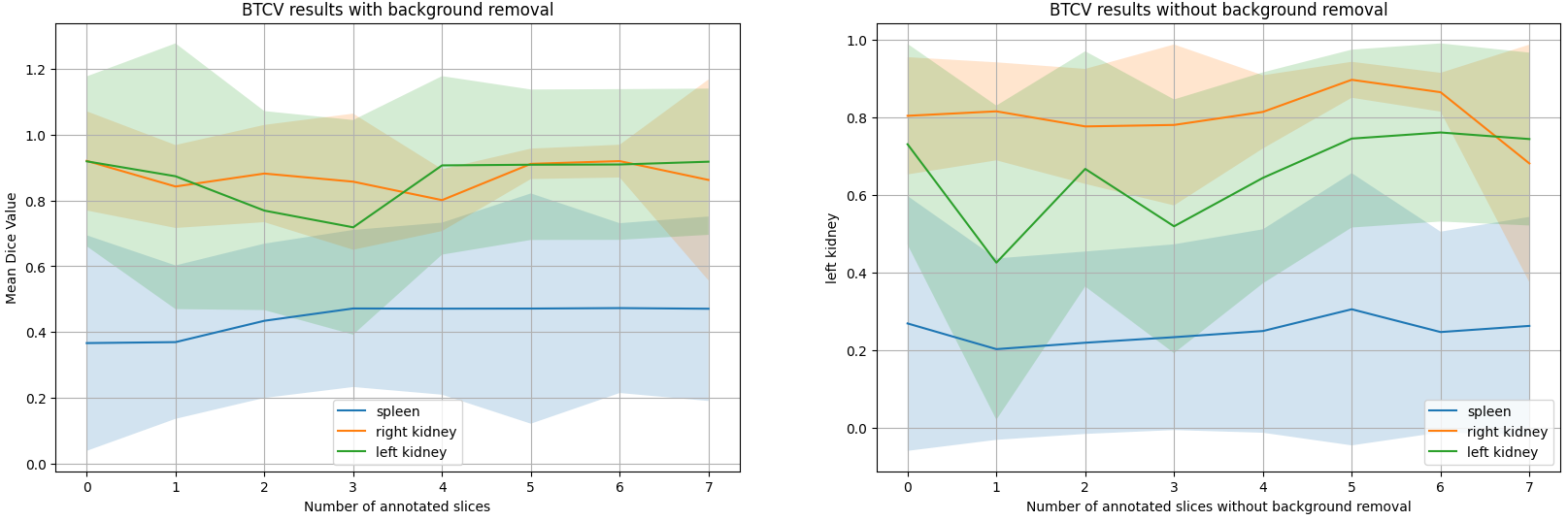} \\
    \includegraphics[width=1\linewidth]{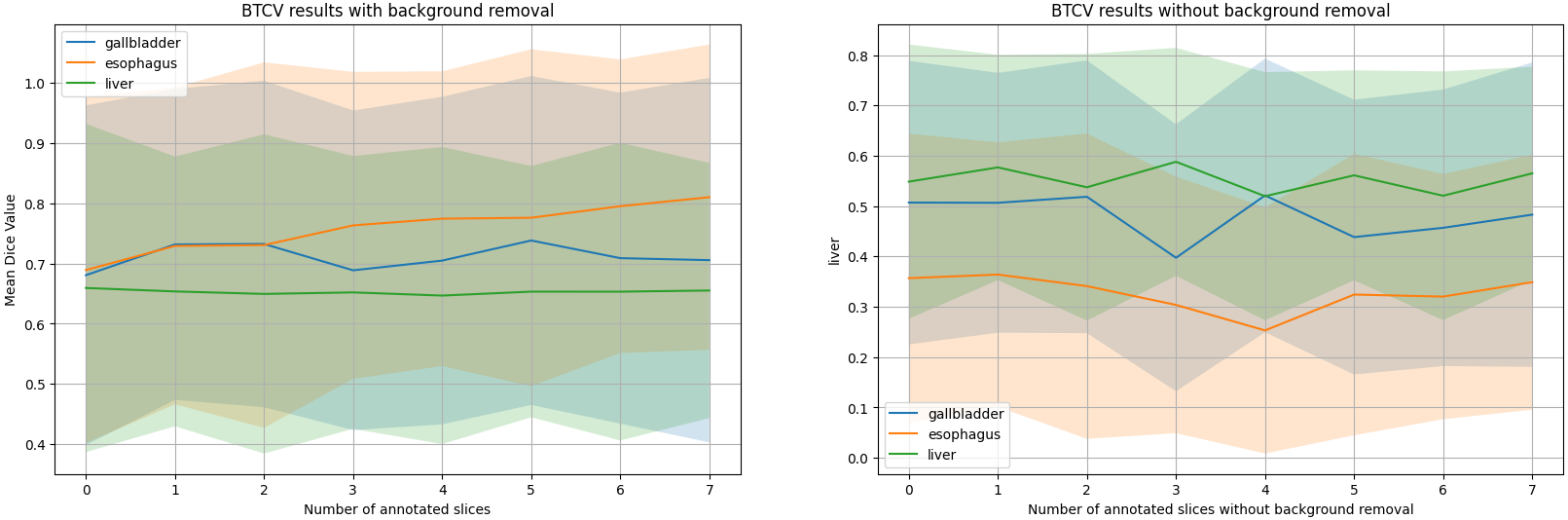} \\
    \includegraphics[width=1\linewidth]{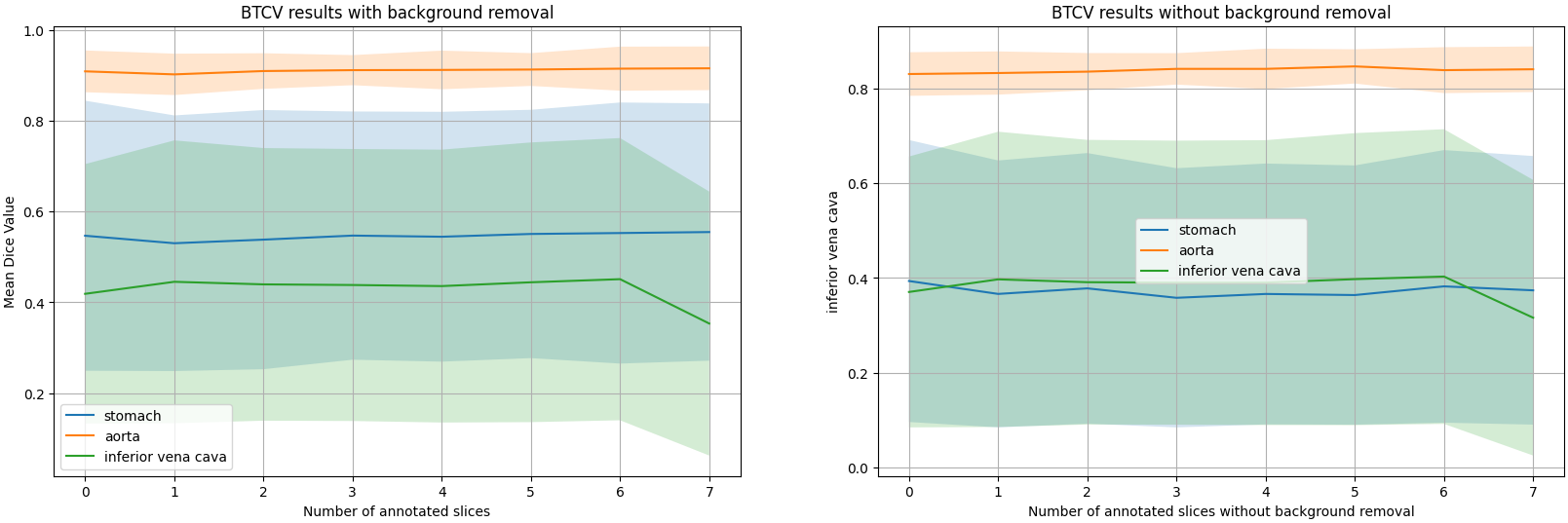} \\
    \includegraphics[width=1\linewidth]{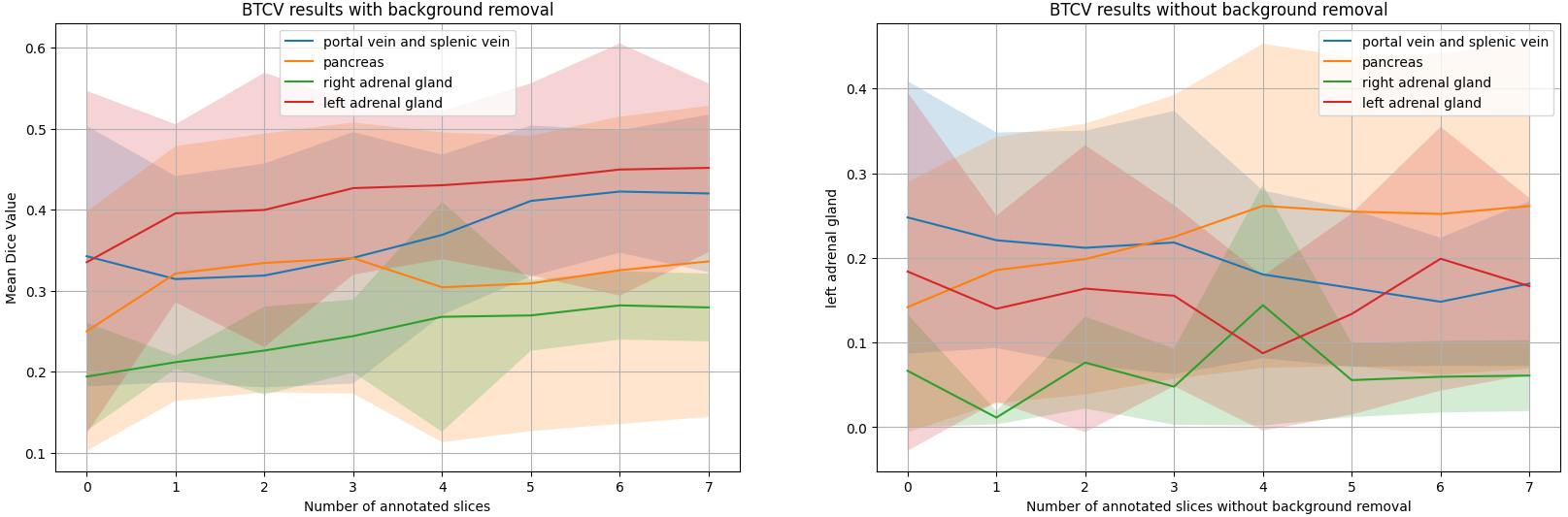}
    \caption{Mean dice scores and 95\% confidence interval with annotated slice numbers on BTCV.}
    \label{fig:plots3}
\end{figure}

We report the results on MSD~\cite{antonelli2022medical} Task03 Liver, Task06 Lung tumor, Task07 Pancreas Tumor, Task09 Spleen, BTCV~\cite{fang2020multi}, and AbdomenCT-1K~\cite{ma2021abdomenct}. Our baseline includes fully supervised automatic models of nnUNet~\cite{nnUNet}, auto3dseg~\cite{cardoso2022monai}, Totalsegmentator~\cite{wasserthal2023totalsegmentator}, and VISTA3D~\cite{he2024vista3d}, which is a recent state-of-the-art model that performs both 3D interactive and automatic segmentation. We plot the Dice score of each dataset in Fig.~\ref{fig:plots1}, Fig.~\ref{fig:plots2}, and Fig.~\ref{fig:plots3}, where the x-axis is the number of annotated frames and y-axis is the average dice score. Each scan has 8 dice scores in those 8 iterations, we use the best dice score and averaged over the whole dataset to report the results in Table.~\ref{t:stats}.

\begin{figure}
    \centering
    \includegraphics[width=1\linewidth]{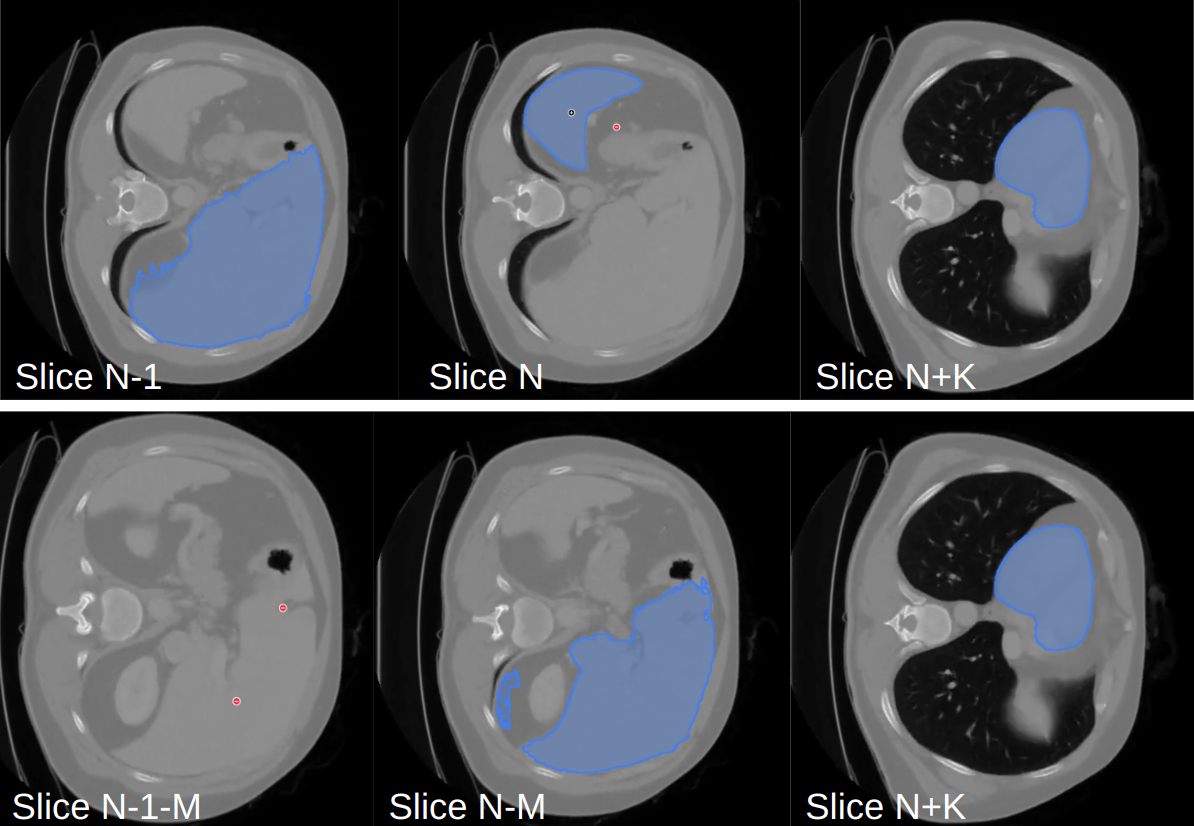}
    \caption{Example of segmenting spleen using SAM2 online-demo. The first line of figures are generated using one whole-volume propagation with one annotated slices~(N), the second line are generated by another propagation with two annotated slices~(N-1-M, N). The video is converted from a full MSD spleen nifti file where each slice is a frame. The initial click is on slice N but on slice N-1, the liver region is segmented, and the segmentation is completely wrong from slice 0 to slice N-1 since liver is segmented. Meanwhile, slice N+K has no spleen and SAM2's tracking started to segment heart. Slice N-1-M is selected to add negative points and all false positives are removed from Slice 0 to Slice N-1-M, however, those negative points have no effect on suppressing false positives starting from its next slice N-M. 
    }
    \label{fig:tracking}
\end{figure}

According to the results, we can draw a few observations:
\begin{itemize}
    \item As shown in the Table~\ref{t:stats}, the performance has a big drop if background slices are not removed. SAM2 is prone to over-tracking foreground objects even they do not exist in those frames. This problem is more common for medical images where organs only appear in certain slices. The overall performance still has a big gap compared with state-of-the-art VISTA3D~\cite{he2024vista3d} where only one click point is needed for 3D segmentation. 
    \item The dice score with respect to the number of annotated slices is not very satisfactory, annotating more slices did not bring sharp performance gains for most organs. 
    \item We empirically found that adding negative points can effectively remove false positives in the current slice, but it cannot suppress false positives in 3D space. As shown in Fig.~\ref{fig:tracking}, the tracking-based segmentation or the way SAM2 is trained to recover occluded objects may contribute to this problem, and might be the reason of these two observations above.
    \item The best performing classes are the lung tumor, pancreatic tumor, right kidney, left kidney, gallbladder, esophagus, and aorta. SAM'2 reaches similar performance with state-of-the-art when background slices are removed. Those organs/tumors are usually smaller single-connected objects, while objects with multiple connected components like hepatic tumors/vessels, and large organs with complicated shapes like pancreas, are harder for SAM2.   
\section{Discussion}
SAM2 has shown impressive zero-shot abilities, but directly applying it for medical imaging problems is not satisfactory and finetuning is needed. Before directly finetune SAM2 model on medical data, we need to think if SAM2's design is really suitable for 3D medical imaging.
3D images are usually processed by a sliding window and cropped into 3D patches, but it can also be processed by a sequence of 2D slices. The winning solutions for 3D medical imaging have mostly been the first one~(e.g. nnUNet, Auto3dSeg), but SAM2 shows a potential that the latter solution can come back and take the lead. It goes back to our three open questions in the beginning of the paper, and more innovation and research is needed to answer them.  

\end{itemize}

\begin{table}[h!]
    \centering
    \label{t:stats}
    \caption{Comparing SAM2's dice scores on held-out test set with state-of-the-art methods. Auto3dSeg and nnUNet are trained for each dataset separately and TotalSegmentor is tested out-of-the-box. VISTA3D is a model that can perform out-box automatic~(VISTA3D auto) and 3D interactive segmentation~(VISTA3D point, 1 click point in each patch), it also supports interactive editing on automatic results~(VISTA3D auto+point).}
    \resizebox{\textwidth}{!}{%
    \begin{tabular}{lccccccc}
                             & \begin{tabular}[c]{@{}c@{}}\textbf{Auto-} \\ \textbf{3dSeg}\end{tabular} & \begin{tabular}[c]{@{}c@{}}\textbf{nn-} \\ \textbf{UNet}\end{tabular} & \begin{tabular}[c]{@{}c@{}}\textbf{TotalSeg-} \\ \textbf{mentator}\end{tabular} & \begin{tabular}[c]{@{}c@{}}\textbf{VISTA3D} \\ \textbf{auto}\end{tabular} & \begin{tabular}[c]{@{}c@{}}\textbf{VISTA3D} \\ \textbf{point}\end{tabular} & \begin{tabular}[c]{@{}c@{}}\textbf{VISTA3D}\\  \textbf{auto + point}\end{tabular} & \begin{tabular}[c]{@{}c@{}}\textbf{SAM2 with/without}\\  \textbf{background removal}\end{tabular}\\
\hline
\multicolumn{7}{l}{\textbf{MSD03 Hepatic Tumor}}                                                                                                                                                                                                                     \\  
liver                        & 0.943     & 0.947  & 0.942            & 0.959                                                   & 0.874                                                    & 0.961         & 0.454/0.373                                  \\
hepatic tumor                & 0.616     & 0.617  & -                & 0.588                                                   & 0.701                                                     & 0.687        & 0.116/0.111                                               \\\hline
\multicolumn{7}{l}{\textbf{MSD06 Lung Tumor}}                                                                                                                                                                                                                        \\
lung tumor                   & 0.562     & 0.554  & -                & 0.614                                                   & 0.682                                                    & 0.719                & 0.812/0.657                                           \\\hline
\multicolumn{7}{l}{\textbf{MSD07 Pancreatic Tumor}}                                                                                                                                                                                                                  \\
pancreas                     & 0.785     & 0.789  & 0.775            & 0.819                                                   & 0.802                                                    & 0.840               &0.295 /0.245                                           \\
pancreatic tumor             & 0.485     & 0.488  & -                & 0.324                                                   & 0.603                                                    & 0.638               &0.498 / 0.229                                           \\\hline
\multicolumn{7}{l}{\textbf{MSD08 Hepatic Tumor}}                                                                                                                                                                                                                     \\
hepatic vessel               & 0.627     & 0.584  & -                & 0.553                                                   & 0.582                                                    & 0.670                     & 0.086/0.079                                      \\
hepatic tumor                & 0.683     & 0.659  & -                & 0.682                                                   & 0.733                                                    & 0.757                     & 0.322/0.265                                 \\\hline
\multicolumn{7}{l}{\textbf{MSD09 Spleen}}                                                                                                                                                                                                                            \\
spleen                       & 0.965     & 0.967  & 0.935            & 0.952                                                   & 0.938                                                    & 0.954                     & 0.898/0.737                                      \\\hline

\multicolumn{7}{l}{\textbf{BTCV-Abdomen}}                                                                                                                                                                                     \\
spleen                       & 0.954     & 0.962  & 0.951            & 0.944                                                   & 0.950                                                    & 0.955                     & 0.475/0.308                                      \\
right kidney                 & 0.936     & 0.951  & 0.941            & 0.943                                                   & 0.937                                                    & 0.945                     & 0.926/0.897                                      \\
left kidney                  & 0.942     & 0.932  & 0.944            & 0.942                                                   & 0.938                                                    & 0.946                     & 0.925/0.786                                      \\
gallbladder                  & 0.663     & 0.771  & 0.739            & 0.794                                                   & 0.792                                                    & 0.807                     & 0.762/0.578                                      \\
esophagus                    & 0.740     & 0.740  & 0.793            & 0.779                                                   & 0.799                                                    & 0.821                     & 0.810/0.495                                      \\
liver                        & 0.964     & 0.961  & 0.970            & 0.967                                                   & 0.715                                                    & 0.969                     & 0.670/0.603                                      \\
stomach                      & 0.876     & 0.797  & 0.946            & 0.944                                                   & 0.938                                                    & 0.946                     & 0.557/0.40                                      \\
aorta                        & 0.929     & 0.909  & 0.929            & 0.931                                                   & 0.925                                                    & 0.932                     & 0.916/0.85                                      \\
inferior vena cava           & 0.834     & 0.827  & 0.854            & 0.842                                                   & 0.729                                                    & 0.856                     & 0.466/0.413                                      \\
portal vein and splenic vein & 0.649     & 0.752  & 0.781            & 0.775                                                   & 0.734                                                    & 0.780                     & 0.448/0.305                                      \\
pancreas                     & 0.759     & 0.820  & 0.807            & 0.841                                                   & 0.797                                                    & 0.853                     & 0.349/0.27                                      \\
right adrenal gland          & 0.604     & 0.661  & 0.696            & 0.692                                                   & 0.673                                                    & 0.699                     & 0.282/0.186                                      \\
left adrenal gland           & 0.638     & 0.642  & 0.643            & 0.646                                                   & 0.666                                                    & 0.660                     & 0.455/0.298                                      \\\hline
\multicolumn{7}{l}{\textbf{AbdomenCT-1K}}                                                                                                                                                                                                                            \\
liver                        & 0.978     & 0.982  & 0.969            & 0.974                                                   & 0.896                                                    & 0.976          & 0.485/0.434                                                 \\
kidney                       & 0.947     & 0.944  & 0.912            & -                                                       & -                                                        & -              & -/-                                                 \\
spleen                       & 0.967     & 0.976  & 0.968            & 0.966                                                   & 0.959                                                    & 0.964          & 0.878/0.81                                                  \\
pancreas                     & 0.857     & 0.860  & 0.828            & 0.865                                                   & 0.853                                                    & 0.881  & 0.392/0.32   \\\hline     
    \end{tabular}%
    }
\end{table}
\bibliographystyle{splncs04}
\bibliography{ml}
\end{document}